\documentclass[runningheads]{llncs}
\usepackage{capt-of}
\usepackage{booktabs}
\usepackage{varwidth}
\usepackage{enumitem}
\usepackage{amsmath,amsfonts,amssymb}
\usepackage{adjustbox}
\usepackage{tabu, colortbl}
\usepackage[dvipsnames]{xcolor}
\usepackage{graphicx}
\usepackage{xr}
\usepackage{acro}
\usepackage[T1]{fontenc}
\usepackage{graphicx}
\usepackage{booktabs}
\usepackage{siunitx}
\usepackage{amsmath}
\usepackage{amssymb}
\usepackage{tabularx}
\usepackage{url}
\usepackage{hyperref}
\usepackage{caption}
\usepackage{subcaption}
\usepackage{csquotes}
\usepackage{nicematrix,tikz}

\NiceMatrixOptions
  {
    custom-line = 
     {
       letter = : ,
       command = dashedline , 
       ccommand = cdashedline ,
       tikz = dashed
     }
  }

\hypersetup{
    colorlinks=true,
    linkcolor=blue,
    filecolor=magenta,    
}
\newcommand{\M}{\mathcal{M}}  

\newcommand{\D}{d}  
\newcommand{\Domain}{\mathcal{D}}
\newcommand{\Range}{\mathcal{R}}

\newcommand{\eps}{\varepsilon}

\begin{document}
%
\title{Enhancing the Utility of Privacy-Preserving 
Cancer Classification using Synthetic Data}
\titlerunning{Enhancing Privacy-Preserving 
Cancer Classification using Synthetic Data}
%
%
%
%
\author{
Richard Osuala\inst{1,2,3}
\and
Daniel M. Lang\inst{2,3} 
\and
Anneliese Riess\inst{2,3} 
\and
Georgios Kaissis\inst{2,3,4} 
\and
Zuzanna Szafranowska\inst{1} 
\and
Grzegorz Skorupko\inst{1} 
\and
Oliver Diaz\inst{1,5}
\and
Julia A. Schnabel\inst{2,3,6}
\and
Karim Lekadir\inst{1,7}
}

\authorrunning{R. Osuala et al.}

\institute{Departament de Matemàtiques i Informàtica, Universitat de Barcelona, Spain
\email{richard.osuala@ub.edu}
\and
Helmholtz Center Munich, Munich, Germany
\and
Technical University of Munich, Munich, Germany
\and
Imperial College London, London, United Kingdom
\and
Computer Vision Center, Bellaterra, Spain
\and
Kings College London, London, UK
\and
Institució Catalana de Recerca i Estudis Avançats (ICREA), Barcelona, Spain
}

\maketitle              
%

\begin{abstract}
Deep learning holds immense promise for aiding radiologists in breast cancer detection. 
However, achieving optimal model performance is hampered by limitations in availability and sharing of data commonly associated to patient privacy concerns.
Such concerns are further exacerbated, as traditional deep learning models can inadvertently leak sensitive training information. 
This work addresses these challenges 
exploring and quantifying the utility of privacy-preserving deep learning techniques, concretely, (i) differentially private stochastic gradient descent (DP-SGD) and (ii) fully synthetic training data generated by our proposed malignancy-conditioned generative adversarial network. 
We assess these methods via downstream malignancy classification of mammography masses using a transformer model. 
Our experimental results depict that synthetic data augmentation can improve privacy-utility tradeoffs in differentially private model training. 
Further, model pretraining on synthetic data achieves remarkable performance, which can be further increased with DP-SGD fine-tuning across all privacy guarantees. 
%
With this first in-depth exploration of privacy-preserving deep learning in breast imaging, we address
current and emerging 
clinical privacy requirements and pave the way towards the adoption of private high-utility 
deep diagnostic models. 
Our reproducible codebase 
is publicly available at \url{https://github.com/RichardObi/mammo_dp}.

\end{abstract}
%
%
%

\begin{figure*}
    \centering
    \includegraphics[width=0.78\textwidth]{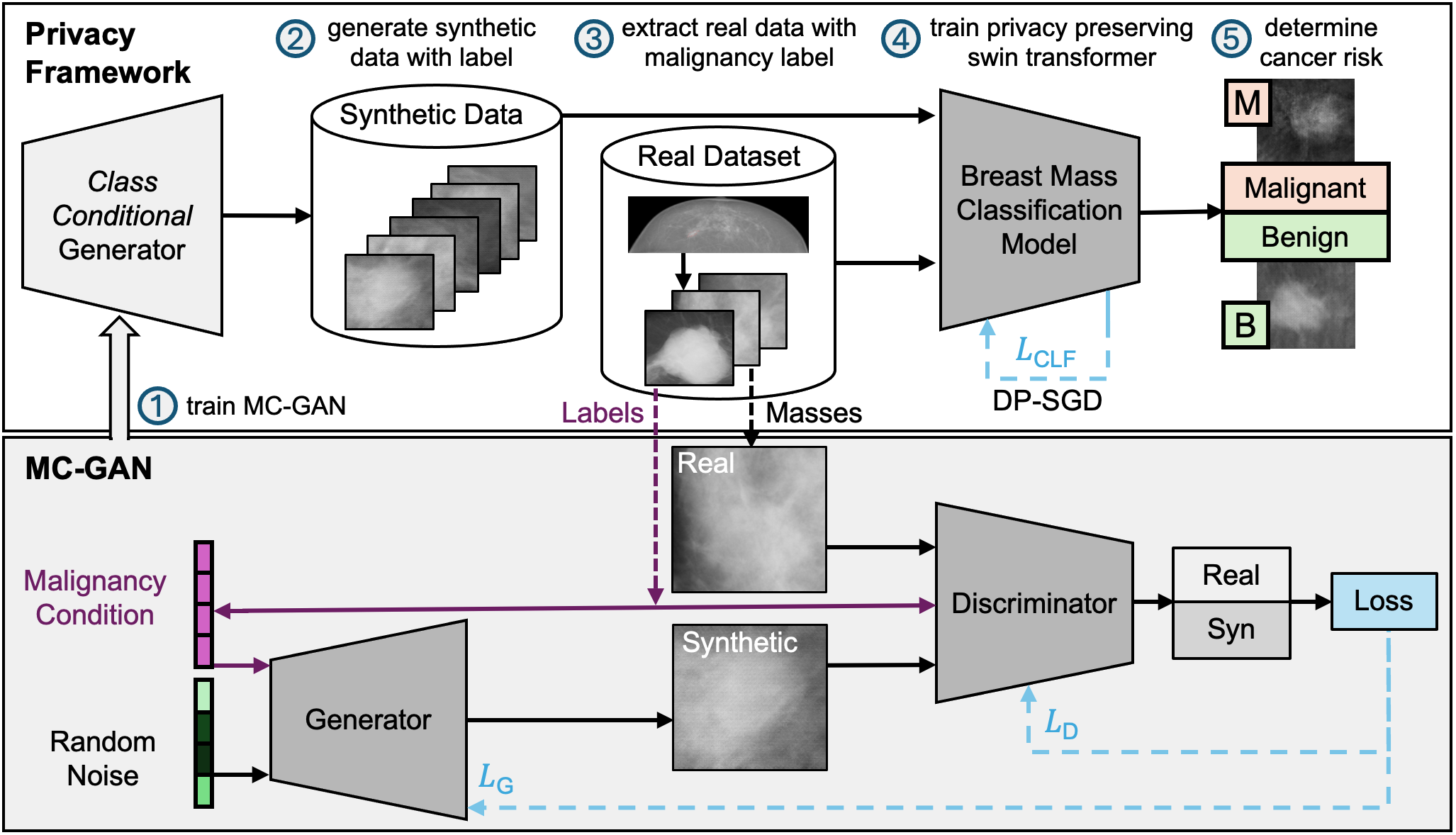}
    \caption{Overview of our privacy-preserving deep learning pipeline and malignancy-conditioned generative adversarial network (MCGAN).}
    \label{fig:overview}
\end{figure*}

\section{Introduction}
Breast cancer accounts for staggering estimates of 684.000 deaths and 2,26 million new cases worldwide per year \cite{GCO2023}.
Part of this burden could be reduced through earlier detection and timely treatment. 
Screening mammography is a cornerstone for early detection and further associated with a reduction in breast cancer mortality 
\cite{mckinney2020international}.
Recent literature emphasizes the potential of deep learning-based computer-aided diagnosis (CAD) \cite{szafranowska2022sharing,osuala2023data,kaissis2021end,mei2022radimagenet}, e.g., demonstrating that a symbiosis of deep learning models with radiologist assessment yields the highest breast cancer detection performances
\cite{mckinney2020international}. 
%
However, training deep learning models on patient data poses a risk of leakage of sensitive person-specific information during and after training
\cite{osuala2023data},
as models have the capacity to memorise sufficient information to allow for high-fidelity image reconstruction \cite{balle2022reconstructing,haim2022reconstructing}.
To avoid such leakage of private patient information, data needs to be protected during model training, in particular when the objective is to develop models to be used in clinical practice or shared among entities.
Furthermore, international data protection regulations grant patients the right to request the removal of their information from data holders. For instance, point (b) of article 17(1) of the EU General Data Protection Regulation (GDPR) \cite{gdpr2018reg} stipulates that data subjects have a \enquote{right to be forgotten}.
Given, for instance, the proven possibility of reconstructing training data given a model's weights \cite{balle2022reconstructing,haim2022reconstructing}, these rights can extend to the removal of patient-specific information from already trained deep learning models \cite{su2022patient}. 
However, it is known to be difficult to \enquote{reliably} and \enquote{provably} remove patient information --- present in only one or few specific training data points --- from already trained model weights \cite{su2022patient}.
A generic and verifiable alternative is given by the removal of a patient's data point from the training data and retraining of the respective model with the reminder of the dataset.
This procedure is not only likely to have negative impacts on the performance of algorithms, but also emerges as a deterrence and risk for hospitals to adopt deep learning models,
due to extensive economic, organisational, and environmental 
costs caused by retraining.
Anticipating patient consent withdrawals, costly retraining can be avoided by demonstrating that deep learning model weights do not include personally identifiable information (PII) about any specific patient.
To this end, a powerful technique to ensure privacy during model training is given by Differentially Private Stochastic Gradient Descent (DP-SGD)\cite{abadi2016deep}, which quantifiably reduces the effect each single training sample can have on the resulting model weights.
Furthermore, privacy-preservation can also be achieved by diagnostic models 
exclusively trained on synthetic data, which is
not (unambiguously) 
attributable to any specific patient but rather contains anonymous samples
representing the essence of the dataset \cite{goodfellow2014generative,osuala2023data}.
%
The caveat of both DP-SGD and synthetic data strategies is, however, that they generally lead to a reduction in model performance, known as the privacy-utility trade-off. 
%
Investigating this trade-off 
in the realm of breast imaging, our core contributions are summarised as follows:
\begin{itemize}  
    \item We design and validate a transformer model, achieving promising performance as a backbone for privacy-preserving breast mass malignancy classification. 
    \item We propose and validate a conditional generative adversarial network capable of differentiating between benign and malignant breast mass generation. 
    \item We empirically quantify privacy-utility-tradeoffs in mass malignancy classification, assessing various differential privacy guarantees, and further combine and compare them with training on synthetic data.
\end{itemize}

\section{Methods and Materials}
\subsection*{Datasets and Preprocessing}
We use the open-access Curated Breast Imaging Subset of Digital Database for Screening Mammography (CBIS-DDSM) dataset \cite{Lee2017}, which consists of 891 scanned film mammography cases with segmented masses with biopsy-proven malignancy status. 
After extracting mass images from craniocaudal view (CC) and mediolateral oblique (MLO) views, we follow the predefined per-patient train-test split \cite{Lee2017}, allocating 1296 mass images for training and 402 (245 benign, 157 malignant) mass images to testing. We further divided this training set randomly per-patient into a training (1104 mass images, 525 malignant) and a validation set (192 mass images, 102 malignant). 
As external test set, we further adopt the publicly available BCDR cohort \cite{lopez2012bcdr}, which comprises 1010 patients, totalling 1493 lesions (639 masses) with biopsy information from both digital mammograms (BCDR-DM) and film mammograms (BCDR-FM). Our final BCDR test set contains 1106 mass images extracted from CC and MLO views, 486 of which are malignant and 620 benign.
%
To obtain mass patches, 
the lesion contour information 
is used to extract bounding boxes from the mammograms. 
We then create a square patch with a minimum size of 128x128 around this bounding box, ensuring a margin of
60 pixel in each direction. 
For classification, the mass patches are resized to pixel dimensions of 224x224 using inter-area interpolation, maintaining image ratios, and stacked to 3 channels.
Models were trained on either a single 8GB NVIDIA RTX 2080 Super or 48GB RTX A6000 GPU using PyTorch 
and opacus \cite{yousefpour2021opacus} for DP-SGD. 

\subsection*{Cancer Classification Transformer Model} \label{sec:clf_model}
Given its 
reported high performance on classifying the presence of a lesion in mammography patches \cite{szafranowska2022sharing} and its shifted window mechanism, allowing to effectively attend to shapes of varying sizes, we adopt a swin transformer (Swin-T) 
\cite{liu2021swin} as cancer classification model,
to distinguish between benign and malignant masses.
We inititalize ImageNet-pretrained \cite{deng2009imagenet} network weights and, after following the Swin-T hyperparameter setup \cite{liu2021swin} (stride, window size), we adjust the last fully-connected layer of the swin transformer reinitializing it with two output nodes each one outputting the logits for one of our respective classes (i.e., malignant or benign).
We only set the parameters of the adjusted fully-connected layer as trainable and apply a learning rate of 1e-5. A weight decay of 1e-8 is used following the fine-tuning experiment described in \cite{liu2021swin}. Furthermore, an adamw optimizer, 
label smoothing of 0.1, and a batch size of 128 are used. During training, random horizontal and vertical flips are applied as data augmentation and a cross entropy loss is backpropagated. Training for 300 epochs, the model from the epoch with the lowest area under the precision-recall curve (AUPRC) on the validation set is selected for testing. 

\subsection*{Malignancy-Conditioned Generative Adversarial Network} \label{sec:MCGAN}
Going beyond unconditional mass synthesis in the literature \cite{szafranowska2022sharing,alyafi2020dcgans}, we propose a malignancy conditioned generative adversarial network (MCGAN) to control the generation of either benign or malignant synthetic breast masses. In general, GANs consist of a generator (G) and a discriminator (D) network, which engage in a two-player zero-sum game, where G generates synthetic samples that D strives to distinguish from real ones \cite{goodfellow2014generative}. 
We design G and D as deep convolutional neural networks \cite{radford2015unsupervised} and, as shown in Fig. \ref{fig:overview}, integrate class-conditional information \cite{mirza2014conditional}. 
To this end, we extract the histopathology report's biopsy information for each mass from the metadata, and convert it into a discrete malignancy label.
Then, we transform this label into a multi-dimensional embedding vector before passing it through a fully-connected layer yielding a representation with the corresponding dimensionality to concatenate it to the generator input (100 dim noise vector) and to the discriminator input (128x128 input image). As D learns to associate class labels with patterns in the input images, it has to learn whether or not a given class corresponds to a given synthetic sample. Furthermore, as the discriminator loss is backpropagated into the generator, G is forced to synthesize samples corresponding to the provided class condition. This results in G learning a conditional distribution based on the value function
\begin{equation*} 
\min_{G} \max_{D} V(D,G) = \min_{G} \max_{D}[\mathbb{E}_{x\sim p_{\mathrm{data}}} [\log D(x|y)] + \mathbb{E}_{z\sim p_{z}} [\log(1 - D(G(z|y)))]].
\end{equation*}
Optimizing the discriminator via binary cross-entropy \cite{goodfellow2014generative}, we define its loss in a class-conditional setup as
\begin{equation*} \label{eq:3}
\begin{aligned}
L_{D_{\mathrm{MCGAN}}} = - \mathbb{E}_{x\sim p_{\mathrm{data}}} [\log D(x|y)] + \mathbb{E}_{z\sim p_{z}} [\log(1 - D(G(z|y)))].
\end{aligned}
\end{equation*}
We train our MCGAN on the CBIS-DDSM training data, applying random horizontal (p=0.5) and vertical (p=0.5) flipping as well as random cropping with resizing, where the resize scale ranges from 0.9 to 1.1 and aspect ratio from 0.95 to 1.1. We further include one-sided label smoothing \cite{radford2015unsupervised} in a range of [0.7, 1.2].
Following \cite{alyafi2020dcgans}, we employ a discriminator convolutional kernel size of 6 and a generator kernel size of 4. We observe that this reduces checkerboard artefacts as D's field-of-view now requires G to create realistic transitions between the kernel-sized patches in the image. MCGAN is trained for 10k epochs with a batch size of 16. Based on the best quality-diversity tradeoff, we select the model from epoch 1.4k after qualitative visual assessment of 
generated samples
.

\subsection*{Patient Privacy Preservation Framework}
%
Privacy protection is an ethical norm and legal obligation, e.g.\@ granting patients the right of their (retrospective) removal from databases \cite{gdpr2018reg}. 
Since (biomedical) deep learning models are vulnerable to information leakage, e.g.\@ sensitive patient attributes \cite{su2022patient,balle2022reconstructing,haim2022reconstructing}, they can be affected by such (and future) regulations.
However, privacy-preserving techniques can be integrated into deep learning frameworks and, to some extent, avoid compromising confidential data.
For instance, (i) model training with DP-SGD \cite{abadi2016deep} or (ii) training exclusively on synthetic data. 

From a legal perspective, models trained on only synthetic data remain unaffected by patient consent withdrawal if \enquote{relatedness} between the data and the data subject cannot be established, or if \enquote{personal data has been rendered synthetic in such a manner that the data subject is no longer identifiable} \cite{lopez2022on} e.g., according to article 4(1) and recital 26 of the GDPR \cite{gdpr2018reg}. 
It is to be noted that in the \enquote{acceptable-risk} legal interpretation, a data subject's re-identification risk is reduced to an \enquote{acceptable} level rather than fully eradicated \cite{lopez2022on}.
Hence, this interpretation enables approaches such as synthetic data and/or Differential Privacy (DP) model training to be used as legally compliant privacy preservation methods despite not guaranteeing a \enquote{zero-risk} of patient re-identification.

DP is a mathematical framework that allows practitioners to provide (worst-case scenario) theoretical privacy guarantees for an individual sharing their data to train a deep learning model.
Consider two databases (e.g., containing image-label pairs), we call them adjacent if they differ in a single data point, i.e., one image is present in one database but not in the other.
Then, a randomised mechanism $\M\colon \Domain\rightarrow\Range$ with domain $\Domain$ and range $\Range$ is said to satisfy $(\eps,\delta)$-differential privacy, if for any two adjacent databases $\D,\D'\in \Domain$ and for any subset of outputs $S\subseteq\Range$, $\Pr[\M(\D)\in S]\leq e^{\eps}\Pr[\M(\D')\in S]+\delta$ holds. 
$\eps$ and $\delta$ bound a single data point's influence on a model's output (e.g. the models' weights or predictions).  
Thus, the smaller the value of these parameters, the higher the model's privacy and the harder it is for an attacker to retrieve information about any training data point.
DP-SGD \cite{abadi2016deep} is the DP variant of the well-known SGD algorithm, and facilitates the training of a model under DP conditions.
In particular, a model trained under $(\eps,\delta)$-DP is robust to post-processing, meaning only using its output for further computations also satisfies $(\eps,\delta)$-DP.
Moreover, the choice of these parameters is application-dependent and normative \cite{de2022unlocking} and varies strongly across real-world deployments \cite{dwork2019differential}.
In the case of mammography, multiple lesions of the same patient are available in the datasets, i.e. one from the CC view and one from the MLO view.
Therefore, to preserve the privacy of one patient it is necessary to protect all their data points (i.e. all images).
In such a case, DP group privacy is used to estimate a patient's DP privacy guarantee. 
However, for simplicity, in our subsequent experiments, we provide image-level privacy guarantees rather than per patient.

\section{Experiments and Results}
%
%
\subsubsection{Synthetic Data Evaluation}

\begin{figure}[htb]
\centering
\begin{minipage}[t]{.45\textwidth}
\resizebox{1\columnwidth}{!}{
\centering
\begin{tabular}[b]{ll|ccc}
    \toprule
    \multicolumn{2}{c}{Experimental Setup} & \multicolumn{3}{c}{Metrics}  \\
    \arrayrulecolor{black} \cmidrule(lr){0-1} \cmidrule(lr){3-5}
    Dataset 1 & Dataset 2 & FID$_{\mathrm{Img}}$ $\downarrow$ & FID$_{\mathrm{Rad}}$ $\downarrow$ & FRD $\downarrow$ \\
    \arrayrulecolor{black} \toprule
    Syn$_{\mathrm{MCGAN}}$  & Real$_{\mathrm{DDSM}}$ & 58.00$\pm$0.72 & 0.81$\pm$.013 & 18.12$\pm$1.01  \\

    Real$_{\mathrm{DDSM}}$ & Real$_{\mathrm{DDSM}}$ & 29.25$\pm$0.82 & 0.31$\pm$.019 &  3.48$\pm$.352 \\

    Syn$_{\mathrm{MCGAN}}$ & Syn$_{\mathrm{MCGAN}}$ & 20.90$\pm$0.16 & 0.32$\pm$.012 &  0.57$\pm$.094 \\

    Real$_{\mathrm{DDSM}}$ & Real$_{\mathrm{BCDR}}$ & 156.43$\pm$14.3 & 3.88$\pm$.351 & 277.63$\pm$39.0  \\

    \arrayrulecolor{black} \bottomrule
\end{tabular}
}
\vfill
\end{minipage}
\begin{minipage}[t]{.54\textwidth}
\centering
\includegraphics[width=1.\linewidth]{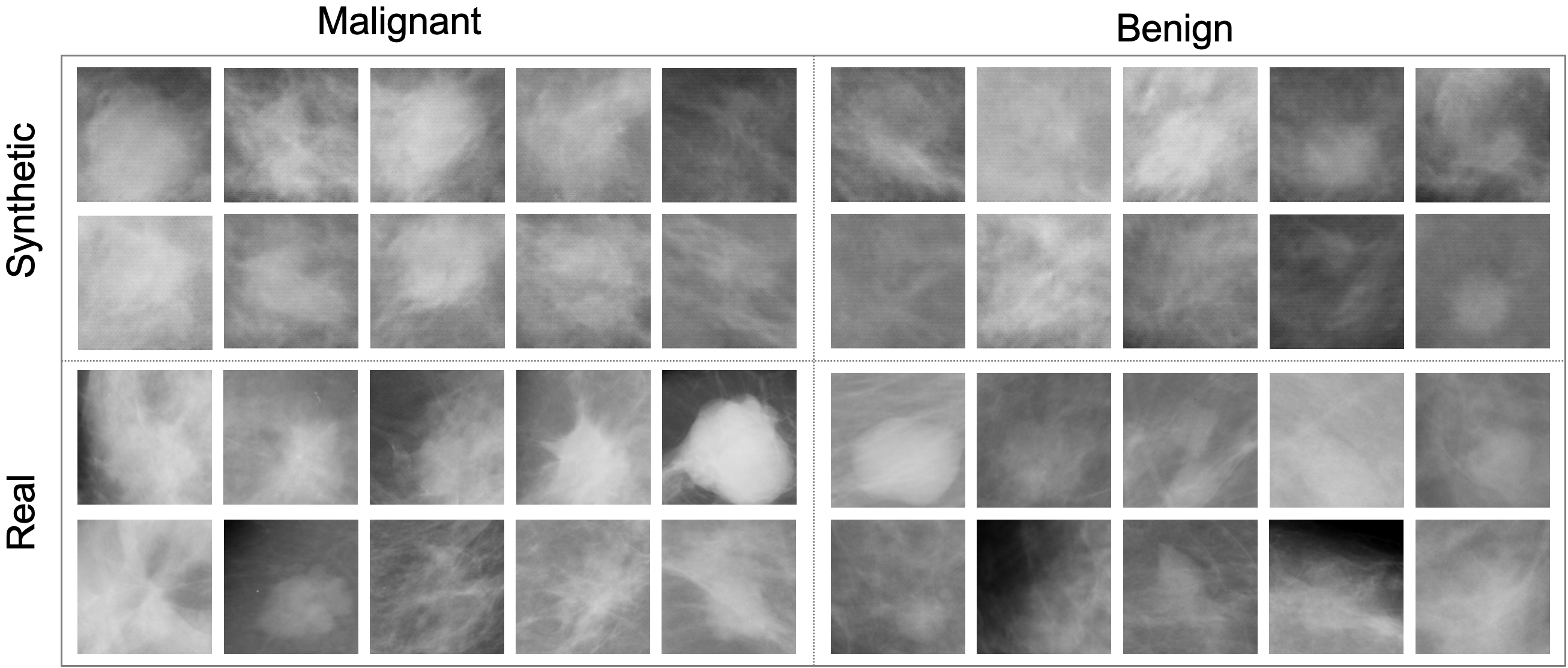} 
\end{minipage}
\caption{Qualitative and quantitative synthesis results: Images are randomly selected malignant and benign real (CBIS-DDSM \cite{Lee2017}) and MCGAN-generated masses. ImageNet \cite{deng2009imagenet} and RadImageNet  \cite{osuala2023medigan,mei2022radimagenet} based FID \cite{heusel2017gans} and FRD \cite{osuala2024towards} scores are reported as mean $\pm$ standard deviation based on 3 subsets randomly sampled per patient (N$_{\mathrm{real}}\approx360$, N$_{\mathrm{syn}}\approx3240$). Row 4 indicates an BCDR-based\cite{lopez2012bcdr} upper bound for comparison with synthetic data metrics in row 1. 
}
\label{fig:syn}
\end{figure}
\begin{table}[ht!]
\centering
\caption{Results for within-domain (CBIS-DDSM \cite{Lee2017}) and out-of-domain (BCDR \cite{lopez2012bcdr}) breast cancer malignancy classification masses extracted from mammograms. \textit{Syn} indicates 3k synthetic images being part of the fine-tuning training data, while \textit{SynPre} represents pretraining all trainable model params with those 3k synthetic images (without DP guarantee), before fine-tuning the last two layers on real data with DP guarantee (\textit{RealFT}). AUROC and AUPRC are reported as mean $\pm$ std based on 3 random seed runs. Best results in bold.}
\resizebox{0.92\columnwidth}{!}{
\begin{tabular}{lcc | cc | cc}
    \toprule
    \multicolumn{3}{c}{Experimental Setup} & \multicolumn{2}{c}{CBIS-DDSM \cite{Lee2017}}  & 
    \multicolumn{2}{c}{BCDR \cite{lopez2012bcdr}} \\
    \arrayrulecolor{black} \cmidrule(lr){0-2} \cmidrule(lr){4-5} \cmidrule(lr){6-7}
    Model & $\eps$ & $\delta$ & AUROC $\uparrow$ & AUPRC $\uparrow$ & AUROC $\uparrow$ & AUPRC $\uparrow$ \\
    \arrayrulecolor{black} \toprule
    SwinT$_{\mathrm{Real}}$ & $\infty$ & $\infty$ & 0.778$\pm$.001 & 	0.85$\pm$.001 & 0.695$\pm$.002 & 0.726$\pm$.003 \\

    \arrayrulecolor{black} \cmidrule(lr){1-7}

    SwinT$_{\mathrm{Syn}}$ & $\infty$ & $\infty$ & 0.597$\pm$.011 &	0.696$\pm$.011 & 0.566$\pm$.064	& 0.602$\pm$.048  \\

    \arrayrulecolor{black} \cmidrule(lr){1-7}
        
    SwinT$_{\mathrm{SynPre}}$ & $\infty$ & $\infty$ & 0.639$\pm$.016 &	0.733$\pm$.001 & 0.622$\pm$.032 & 0.660$\pm$.017 \\

    \arrayrulecolor{black} \cmidrule(lr){1-7}

    SwinT$_{\mathrm{Real}}$ & $1$ & $1e^{-4}$ & 0.525$\pm$.043	& 0.640$\pm$.030 & 0.487$\pm$.020 & 0.549$\pm$.020 \\

    SwinT$_{\mathrm{Real+Syn}}$ & $1$ & $1e^{-4}$ & \textbf{0.553$\pm$.040} &	\textbf{0.665$\pm$.025} & \textbf{0.521$\pm$.023} &	\textbf{0.573$\pm$.024} \\

    \arrayrulecolor{gray} \cmidrule(lr){4-7}
    
    SwinT$_{\mathrm{SynPre+RealFT}}$ & $\infty$|$1$ & $\infty$|$1e^{-4}$ & 0.661$\pm$.018 & 0.741$\pm$.007 & 0.637$\pm$.026	 & 0.67$\pm$0013 \\

    \arrayrulecolor{black} \cmidrule(lr){1-7}

    SwinT$_{\mathrm{Real}}$ & $6$ & $1e^{-4}$ & 0.572$\pm$.031 & 0.679$\pm$.019 & 0.532$\pm$.031 & 0.579$\pm$.029\\

    SwinT$_{\mathrm{Real+Syn}}$ & $6$ & $1e^{-4}$ & \textbf{0.617$\pm$.013} &	\textbf{0.708$\pm$.015} & \textbf{0.609$\pm$.027} & \textbf{0.647$\pm$.024} \\

    \arrayrulecolor{gray} \cmidrule(lr){4-7}
    
    SwinT$_{\mathrm{SynPre+RealFT}}$ & $\infty$|$6$ & $\infty$|$1e^{-4}$ & 0.677$\pm$.014 & 0.752$\pm$.009 & 0.647$\pm$.022 & 	0.679$\pm$.009 \\
    
    \arrayrulecolor{black} \cmidrule(lr){1-7}
    
    SwinT$_{\mathrm{Real}}$ & $12$ & $1e^{-4}$ & 0.596$\pm$.023	& 0.702$\pm$.013 & 0.559$\pm$.033 &	0.600$\pm$.030\\

    SwinT$_{\mathrm{Real+Syn}}$ & $12$ & $1e^{-4}$ & \textbf{0.624$\pm$.010} &	\textbf{0.704$\pm$.012} & \textbf{0.625$\pm$.020}	& \textbf{0.663$\pm$.012} \\

    \arrayrulecolor{gray} \cmidrule(lr){4-7}

    SwinT$_{\mathrm{SynPre+RealFT}}$ & $\infty$|$12$ & $\infty$|$1e^{-4}$ & 0.688$\pm$.012 &	0.758$\pm$.011 & 0.654$\pm$.019 &	0.685$\pm$.007 \\

    \arrayrulecolor{black} \cmidrule(lr){1-7}
    
    SwinT$_{\mathrm{Real}}$ & $20$ & $1e^{-4}$ & 0.611$\pm$.018	&
    \textbf{0.715$\pm$.012} &
    0.581$\pm$.028 & 0.618$\pm$.026 \\

    SwinT$_{\mathrm{Real+Syn}}$ & $20$ & $1e^{-4}$ & \textbf{0.630$\pm$.003} &	0.699$\pm$.008 & \textbf{0.641$\pm$.018}	& \textbf{0.685$\pm$.012} \\

    \arrayrulecolor{gray} \cmidrule(lr){4-7}

    SwinT$_{\mathrm{SynPre+RealFT}}$ & $\infty$|$20$ & $\infty$|$1e^{-4}$ & 0.697$\pm$.012	& 0.763$\pm$.012 & 0.659$\pm$.017	& 0.689$\pm$.006 \\

    \arrayrulecolor{black} \cmidrule(lr){1-7}
        
    SwinT$_{\mathrm{Real}}$ & $60$ & $1e^{-4}$ & 0.622$\pm$.014 &
    \textbf{0.721$\pm$.110} & 0.605$\pm$.019 & 	0.640$\pm$.017 \\   

    SwinT$_{\mathrm{Real+Syn}}$ & $60$ & $1e^{-4}$ & \textbf{0.629$\pm$.002} &	0.694$\pm$.005 & \textbf{0.650$\pm$.013}	& \textbf{0.696$\pm$.007} \\ 

    \arrayrulecolor{gray} \cmidrule(lr){4-7}

    SwinT$_{\mathrm{SynPre+RealFT}}$ & $\infty$|$60$ & $\infty$|$1e^{-4}$ & 0.712$\pm$.013 & 0.776$\pm$.013 & 0.671$\pm$.014	& 0.697$\pm$.004 \\
    
    \arrayrulecolor{black} \bottomrule
    
\end{tabular}
}
\label{tab:1}
\end{table}

Qualitatively assessing the synthetic images in Fig. \ref{fig:syn}, it is not readily possible to distinguish synthetic from real masses in terms of image fidelity or diversity. We note the absence of clear visual indicators to distinguish between malignant and benign images for both real and synthetic images. This is in line with the difficulty of determining the malignancy of a mammographic lesion shown by high clinical error rates and inter-observer variability \cite{ekpo2018errors}. However, results for training our malignancy classification model on only synthetic data (see \textit{Syn} and \textit{SynPre} in Table \ref{tab:1}) show that the synthetic data captures the conditional distribution effectively generating either malignant or benign masses.
Both, vanilla ImageNet-based Fréchet Inception Distance (FID) \cite{heusel2017gans,deng2009imagenet} and radiology domain-specific RadImageNet-based FID \cite{osuala2023medigan,mei2022radimagenet}, concur that the synthetic data (FID$_{\mathrm{Img}}$=58$\pm$.72) is substantially closer to the real CBIS-DDSM \cite{Lee2017} distribution compared to BCDR \cite{lopez2012bcdr} (FID$_{Img}$=156.43$\pm$1.43). This is even more pronounced when comparing the variation of extracted radiomics features for CBIS-DDSM to synthetic (FRD=18.12) and BCDR (FRD=277.63) images using the Fréchet Radiomics Distance (FRD) \cite{osuala2024towards}.
While this indicates desirable synthetic data fidelity, we also observe good diversity. The latter is shown by comparing subsets of the same datasets with each other, where the variation within the synthetic data (e.g., FID$_{\mathrm{Rad}}$=0.32$\pm$.12) closely resembles the variation within the real CBIS-DDSM dataset (e.g., FID$_{\mathrm{Rad}}$=0.31$\pm$.19). Notwithstanding less variation in radiomics imaging biomarkers within the synthetic data (FRD$_{\mathrm{Syn}}$=0.57 vs. FRD$_{\mathrm{Real}}$=3.48), this overall points to a valid coverage of the distribution and an absence of mode collapse.

\subsubsection{Mass Malignancy Classification}
As shown in Table \ref{tab:1}, we conduct experiments with and without formal privacy guarantees. For scenarios where a formal privacy guarantee is not strictly required and, thus, synthetic data suffices as privacy mechanism, we compare the results of training SwinT on synthetic data (\textit{Syn}) and on real data (\textit{Real}) with DP-SGD.
\textit{Kaissis et al.} \cite{kaissis2021end} defined $\eps=6$ as suitable privacy budget for their medical imaging dataset. Compared to DP-SGD with $\eps=6$, synthetic data achieves better AUPRCs for within-domain tests on CBIS-DDSM (SwinT$_{\mathrm{Syn}}$=0.696 vs SwinT$_{\mathrm{Real(\eps=6)}}$=0.679) and is on par for out-of-domain (ood) tests on BCDR (SwinT$_{\mathrm{Syn}}$=0.602 vs SwinT$_{\mathrm{Real(\eps=6)}}$=0.600). However, training all SwinT layers using synthetic data (\textit{SynPre}), achieves substantially better performance only approximated by DP results for $\eps=60$ for within-domain (SwinT$_{\mathrm{SynPre}}$=0.733 vs SwinT$_{\mathrm{Real(\eps=60)}}$=0.721) and ood (SwinT$_{\mathrm{SynPre}}$ =0.66 vs SwinT$_{\mathrm{Real(\eps=60)}}$=0.64) tests. Further fine-tuning SwinT$_{\mathrm{SynPre}}$ on real data using DP-SGD results in additional improvement across all privacy parameters for within-domain and ood testing. For instance, training SwinT$_{\mathrm{SynPre+RealFT}}$ with $\eps=1$ results in an AUPRC of 0.74 and 0.67 for CBIS-DDSM and BCDR, respectively.
To assess scenarios where a formal guarantee is required, we further compare DP-SGD training of SwinT on real data (\textit{Real}) with DP-SGD training on a mix of real and synthetic data (\textit{Real+Syn}). To this end, our experiments show that such synthetic data augmentation can improve the privacy-utility tradeoff. This is exemplified by SwinT$_{\mathrm{Real+Syn(\eps=6)}}$ accomplishing an AUPRC of 0.708 within-domain and 0.647 ood, while SwinT$_{\mathrm{Real(\eps=6)}}$ achieved 0.679 and 0.579, respectively. We further observe the trend that stricter privacy budgets (i.e., smaller $\eps$) can be associated with more added performance of synthetic data as additional classification model training data.

\section{Discussion and Conclusion}
We introduce a privacy preservation framework based on differential privacy (DP) and synthetic data and apply it to the diagnostic task of classifying the malignancy of breast masses extracted from screening mammograms.
We further propose, train, and evaluate a malignancy-conditioned generative adversarial network to generate a dataset of benign and malignant synthetic breast masses.
Next, we train a swin transformer model on mass malignancy classification and assess, compare and combine training under DP and training on synthetic data. This analysis revealed that when training with DP, synthetic data augmentation can notably improve classification performance for within-domain and out-of-domain test cases. Apart from that, we show, across privacy mechanisms and across domains, that the performance of models pretrained on synthetic data can be further improved by DP fine-tuning on real data. 

This finding is particularly important considering that synthetic data, if not directly attributable to any specific patient, can become a valid, legally compliant alternative to strict DP guarantees in clinical practice.
Consequently, it is to be further investigated where and when deterministic mechanisms without formal DP guarantees can suffice to shield against different privacy attacks \cite{cohen2020towards}. In particular, we motivate future work to analyse the extent to which the inherent properties of synthetic data generation algorithms can provide empirical protection against attacks.
A methodological alternative to our approach is to assess privacy-utility tradeoffs when training the generative model itself using DP-SGD \cite{ghalebikesabi2023differentially,osuala2023data}, resulting in formal privacy guarantees of the generated synthetic datasets. Thus, a further avenue to explore then lies within the question whether randomness inherent in randomised data synthesis algorithms (e.g., based on the noise in diffusion models \cite{sohl2015deep} or GANs \cite{goodfellow2014generative}) can be used to amplify the privacy of the DP versions of such synthesis algorithms, thereby potentially further enhancing privacy-utility tradeoffs.
To this end, our study constitutes a crucial first step leading towards the clinical adoption of diagnostic deep learning models, enabling practical privacy-utility tradeoffs all while anticipating respective legal obligations and clinical requirements. 
\begin{credits}
\subsubsection{\ackname} 
This study has received funding from the European Union’s Horizon 
research and innovation programme under grant agreement No 952103 (EuCanImage) and No 101057699 (RadioVal). It was further partially supported by the project FUTURE-ES (PID2021-126724OB-I00) from the Ministry of Science and Innovation of Spain. RO acknowledges a research stay grant from the Helmholtz Information and Data Science Academy (HIDA).
%
%
%
\subsubsection{\discintname}
The authors have no competing interests to declare that are relevant to the content of this article.
\end{credits}

\bibliographystyle{splncs04}
\bibliography{references}

\begin{thebibliography}{10}
\providecommand{\url}[1]{\texttt{#1}}
\providecommand{\urlprefix}{URL }
\providecommand{\doi}[1]{https://doi.org/#1}

\bibitem{abadi2016deep}
Abadi, M., Chu, A., Goodfellow, I., McMahan, H.B., Mironov, I., Talwar, K., Zhang, L.: Deep learning with differential privacy. In: Proceedings of the 2016 ACM SIGSAC conference on computer and communications security. pp. 308--318 (2016)

\bibitem{alyafi2020dcgans}
Alyafi, B., Diaz, O., Marti, R.: {DCGANs for realistic breast mass augmentation in x-ray mammography}. In: Medical Imaging 2020: Computer-Aided Diagnosis. vol. 11314, p. 1131420. International Society for Optics and Photonics (2020)

\bibitem{balle2022reconstructing}
Balle, B., Cherubin, G., Hayes, J.: Reconstructing training data with informed adversaries. In: 2022 IEEE Symposium on Security and Privacy (SP). pp. 1138--1156. IEEE (2022)

\bibitem{cohen2020towards}
Cohen, A., Nissim, K.: Towards formalizing the gdpr’s notion of singling out. Proceedings of the National Academy of Sciences  \textbf{117}(15),  8344--8352 (2020)

\bibitem{de2022unlocking}
De, S., Berrada, L., Hayes, J., Smith, S.L., Balle, B.: Unlocking high-accuracy differentially private image classification through scale. arXiv preprint arXiv:2204.13650  (2022)

\bibitem{deng2009imagenet}
Deng, J., Dong, W., Socher, R., Li, L.J., Li, K., Fei-Fei, L.: Imagenet: A large-scale hierarchical image database. In: 2009 IEEE conference on computer vision and pattern recognition. pp. 248--255. Ieee (2009)

\bibitem{dwork2019differential}
Dwork, C., Kohli, N., Mulligan, D.: Differential privacy in practice: Expose your epsilons! Journal of Privacy and Confidentiality  \textbf{9}(2) (2019)

\bibitem{ekpo2018errors}
Ekpo, E.U., Alakhras, M., Brennan, P.: Errors in mammography cannot be solved through technology alone. Asian Pacific journal of cancer prevention: APJCP  \textbf{19}(2), ~291 (2018)

\bibitem{gdpr2018reg}
{European Parliament and Council of European Union}: {General Data Protection Regulation (GDPR), REGULATION (EU) 2016/679 OF THE EUROPEAN PARLIAMENT AND OF THE COUNCIL}. Online at https://eur-lex.europa.eu/legal-content/EN/TXT/HTML/?uri=CELEX:32016R0679/ (2018)

\bibitem{ghalebikesabi2023differentially}
Ghalebikesabi, S., Berrada, L., Gowal, S., Ktena, I., Stanforth, R., Hayes, J., De, S., Smith, S.L., Wiles, O., Balle, B.: Differentially private diffusion models generate useful synthetic images. arXiv preprint arXiv:2302.13861  (2023)

\bibitem{GCO2023}
{Global Cancer Observatory}: The global cancer observatory (gco) is an interactive web-based platform presenting global cancer statistics to inform cancer control and research. \url{https://gco.iarc.fr/} (2023), accessed on 2023-01-17

\bibitem{goodfellow2014generative}
Goodfellow, I., Pouget-Abadie, J., Mirza, M., Xu, B., Warde-Farley, D., Ozair, S., Courville, A., Bengio, Y.: Generative adversarial nets. In: Advances in neural information processing systems. pp. 2672--2680 (2014)

\bibitem{haim2022reconstructing}
Haim, N., Vardi, G., Yehudai, G., Shamir, O., Irani, M.: Reconstructing training data from trained neural networks. arXiv preprint arXiv:2206.07758  (2022)

\bibitem{heusel2017gans}
Heusel, M., Ramsauer, H., Unterthiner, T., Nessler, B., Hochreiter, S.: {GANs trained by a two time-scale update rule converge to a local nash equilibrium}. arXiv preprint arXiv:1706.08500  (2017)

\bibitem{kaissis2021end}
Kaissis, G., Ziller, A., Passerat-Palmbach, J., Ryffel, T., Usynin, D., Trask, A., Lima~Jr, I., Mancuso, J., Jungmann, F., Steinborn, M.M., et~al.: End-to-end privacy preserving deep learning on multi-institutional medical imaging. Nature Machine Intelligence  \textbf{3}(6),  473--484 (2021)

\bibitem{Lee2017}
Lee, R.S., Gimenez, F., Hoogi, A., Miyake, K.K., Gorovoy, M., Rubin, D.L.: A curated mammography data set for use in computer-aided detection and diagnosis research. Scientific data  \textbf{4}(1), ~1--9 (2017)

\bibitem{liu2021swin}
Liu, Z., Lin, Y., Cao, Y., Hu, H., Wei, Y., Zhang, Z., Lin, S., Guo, B.: {Swin transformer: Hierarchical vision transformer using shifted windows}. In: Proceedings of the IEEE/CVF international conference on computer vision. pp. 10012--10022 (2021)

\bibitem{lopez2022on}
L{\'o}pez, C.A.F.: On the legal nature of synthetic data. In: NeurIPS 2022 Workshop on Synthetic Data for Empowering ML Research (2022)

\bibitem{lopez2012bcdr}
Lopez, M.G., Posada, N., Moura, D.C., Poll{\'a}n, R.R., Valiente, J.M.F., Ortega, C.S., Solar, M., Diaz-Herrero, G., Ramos, I., Loureiro, J., et~al.: {BCDR: a breast cancer digital repository}. In: 15th International conference on experimental mechanics. vol.~1215 (2012)

\bibitem{mckinney2020international}
McKinney, S.M., Sieniek, M., Godbole, V., Godwin, J., Antropova, N., Ashrafian, H., Back, T., Chesus, M., Corrado, G.S., Darzi, A., et~al.: International evaluation of an ai system for breast cancer screening. Nature  \textbf{577}(7788),  89--94 (2020)

\bibitem{mei2022radimagenet}
Mei, X., Liu, Z., Robson, P.M., Marinelli, B., Huang, M., Doshi, A., Jacobi, A., Cao, C., Link, K.E., Yang, T., et~al.: {RadImageNet: An Open Radiologic Deep Learning Research Dataset for Effective Transfer Learning}. Radiology: Artificial Intelligence p. e210315 (2022)

\bibitem{mirza2014conditional}
Mirza, M., Osindero, S.: Conditional generative adversarial nets. arXiv preprint arXiv:1411.1784  (2014)

\bibitem{osuala2023data}
Osuala, R., Kushibar, K., Garrucho, L., Linardos, A., Szafranowska, Z., Klein, S., Glocker, B., Diaz, O., Lekadir, K.: Data synthesis and adversarial networks: A review and meta-analysis in cancer imaging. Medical Image Analysis  \textbf{84},  102704 (2023)

\bibitem{osuala2024towards}
Osuala, R., Lang, D., Verma, P., Joshi, S., Tsirikoglou, A., Skorupko, G., Kushibar, K., Garrucho, L., Pinaya, W.H., Diaz, O., et~al.: Towards learning contrast kinetics with multi-condition latent diffusion models. arXiv preprint arXiv:2403.13890  (2024)

\bibitem{osuala2023medigan}
Osuala, R., Skorupko, G., Lazrak, N., Garrucho, L., Garc{\'\i}a, E., Joshi, S., Jouide, S., Rutherford, M., Prior, F., Kushibar, K., et~al.: medigan: a python library of pretrained generative models for medical image synthesis. Journal of Medical Imaging  \textbf{10}(6),  061403 (2023)

\bibitem{radford2015unsupervised}
Radford, A., Metz, L., Chintala, S.: Unsupervised representation learning with deep convolutional generative adversarial networks. arXiv preprint arXiv:1511.06434  (2015)

\bibitem{sohl2015deep}
Sohl-Dickstein, J., Weiss, E., Maheswaranathan, N., Ganguli, S.: Deep unsupervised learning using nonequilibrium thermodynamics. In: International Conference on Machine Learning. pp. 2256--2265. PMLR (2015)

\bibitem{su2022patient}
Su, R., Liu, X., Tsaftaris, S.A.: Why patient data cannot be easily forgotten? In: Medical Image Computing and Computer Assisted Intervention--MICCAI 2022: 25th International Conference, Singapore, September 18--22, 2022, Proceedings, Part VIII. pp. 632--641. Springer (2022)

\bibitem{szafranowska2022sharing}
Szafranowska, Z., Osuala, R., Breier, B., Kushibar, K., Lekadir, K., Diaz, O.: Sharing generative models instead of private data: a simulation study on mammography patch classification. In: 16th International Workshop on Breast Imaging (IWBI2022). vol. 12286, pp. 169--177. SPIE (2022)

\bibitem{yousefpour2021opacus}
Yousefpour, A., Shilov, I., Sablayrolles, A., Testuggine, D., Prasad, K., Malek, M., Mironov, I.: Opacus: User-friendly differential privacy library in pytorch. arXiv preprint arXiv:2109.12298  (2021)

\end{thebibliography}

\end{document}